\documentclass[preprint, 10pt]{elsarticle}
\usepackage{hyperref}
\usepackage{amssymb}
\usepackage{amsmath}
\usepackage{multirow}
\usepackage{booktabs}
\usepackage{arydshln}
\usepackage{subcaption}
\usepackage{hyperref}
\usepackage{subfloat}
\usepackage[T1]{fontenc}
\usepackage{amsfonts}
\usepackage{algorithm}
\usepackage{algorithmic}

\usepackage{amsmath}
\usepackage{units}
\usepackage{cleveref}
\Crefname{figure}{Fig.}{Figures}
\begin{document}

\begin{frontmatter}

\title{Layer-Wise Feature Metric of Semantic-Pixel Matching for Few-Shot Learning}
\author[1]{Hao Tang\fnref{equal}}
\ead{halo._Tang@outlook.com}
\author[1]{Junhao Lu\fnref{equal}}
\ead{polynya-code@outlook.com}
\author[1]{Guoheng Huang\corref{corresponding}}
\ead{kevinwong@gdut.edu.cn}
\author[3,4]{Ming Li}
\ead{mingli@zjnu.edu.cn}
\author[2]{Xuhang Chen}
\ead{bcbillycat@qq.com}
\author[5]{Guo Zhong}
\ead{yb77410@umac.mo}
\author[1]{Zhengguang Tan}
\ead{bcbillycat@qq.com}
\author[6]{Zinuo Li}
\ead{zinuo.li@research.uwa.edu.au}

\cortext[corresponding]{Corresponding author.}
\fntext[equal]{The two authors contributed equally to this work.}
\address[1]{School of Computer Science and Technology, Guangdong University of Technology, Guangdong, China}
\address[2]{School of Computer Science and Engineering, Huizhou University, Huizhou, China}
\address[3]{Zhejiang Institute of Optoelectronics, Jinhua, China}
\address[4]{Zhejiang Key Laboratory of Intelligent Education Technology and Application, Jinhua, China}
\address[5]{School of Information Science and Technology, Guangdong University of Foreign Studies, Guangzhou, China}
\address[6]{School of Physics, Maths and Computing, University of Western Australia, Western Australia, Australia}

\begin{abstract}
In Few-Shot Learning (FSL), traditional metric-based approaches often rely on global metrics to compute similarity. However, in natural scenes, the spatial arrangement of key instances is often inconsistent across images. This spatial misalignment can result in mismatched semantic pixels, leading to inaccurate similarity measurements. To address this issue, we propose a novel method called the Layer-Wise Features Metric of Semantic-Pixel Matching (LWFM-SPM) to make finer comparisons. Our method enhances model performance through two key modules: (1) the Layer-Wise Embedding (LWE) Module, which refines the cross-correlation of image pairs to generate well-focused feature maps for each layer; (2)the Semantic-Pixel Matching (SPM) Module, which aligns critical pixels based on semantic embeddings using an assignment algorithm. We conducted extensive experiments to evaluate our method on four widely used few-shot classification benchmarks: miniImageNet, tieredImageNet, CUB-200-2011, and CIFAR-FS. The results indicate that LWFM-SPM achieves competitive performance across these benchmarks. Our code will be publicly available on https://github.com/Halo2Tang/Code-for-LWFM-SPM.
\end{abstract}

%% Keywords

\end{frontmatter}

% \linenumbers

\section{Introduction}
Humans have the ability to abstract and generalize low-level visual elements, such as contours, edges, colors, textures, and shapes, to form high-level semantic features that aid in recognizing and understanding the similarities and differences between objects. This capability is particularly crucial in few-shot classification tasks, as it allows models to accurately identify and distinguish between different categories based on contrasting critical high-level semantic features, even when faced with a limited number of samples from new categories. In contrast, traditional deep learning methods~\cite{a27,a28} typically rely on large amounts of labeled data for training in order to recognize and classify specific objects or concepts. In few-shot learning scenarios, these models may encounter challenges, as they are not specifically designed to learn from a small amount of data quickly. Recently, few-shot learning methods have been introduced to address this limitation, typically requiring only a few images to understand the characteristics of a class and generalize these features to unseen images for inductive reasoning. Among these methods, metric-based approaches~\cite{a29,a20,a21} are computationally efficient, as they do not require extensive parameter tuning or complex model structures. They rely on learning image embeddings to measure the similarity between objects and perform classification.

\begin{figure}[!t]
\centerline{\includegraphics[width=0.5\linewidth]{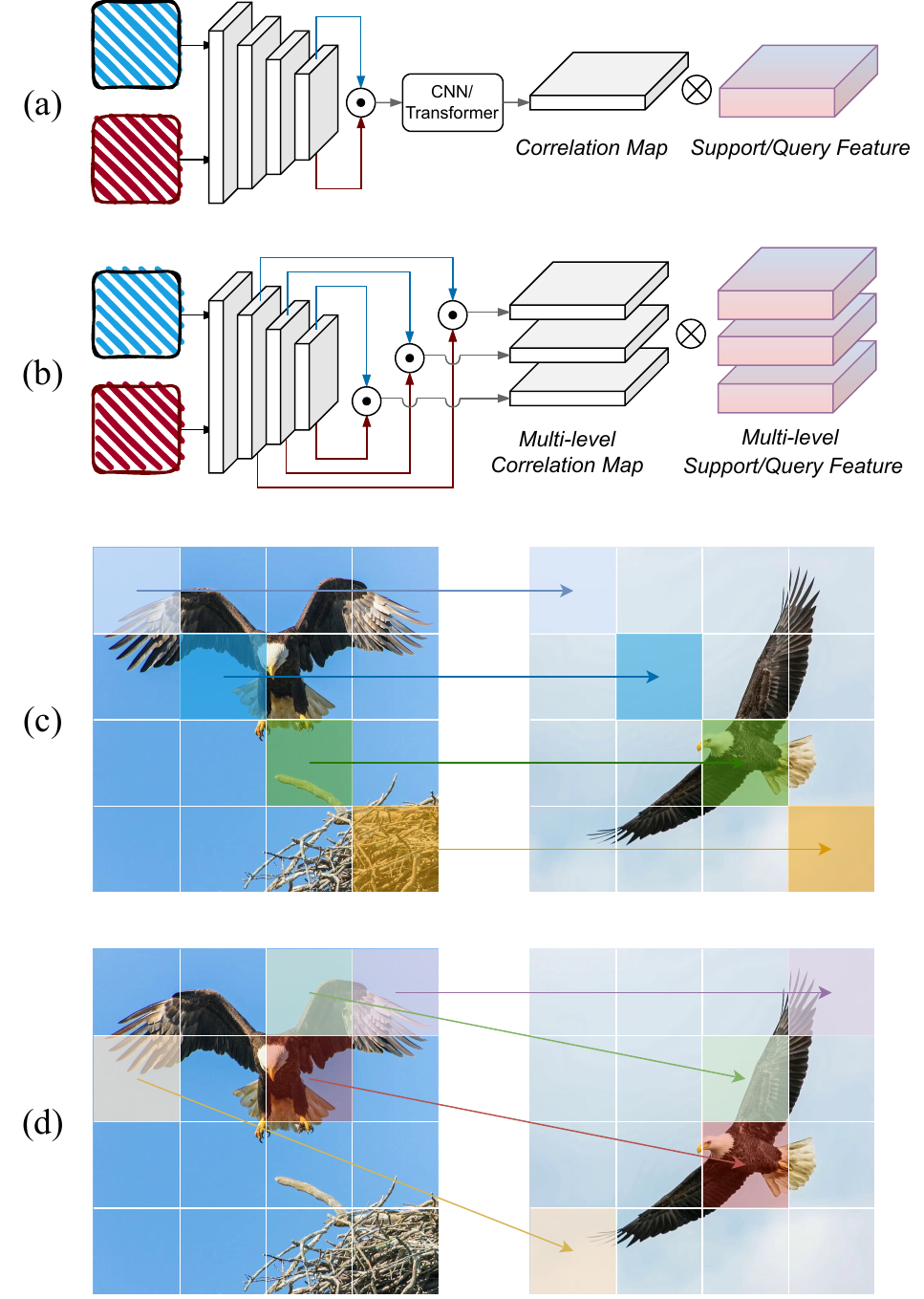}}
\caption{The key difference between our method and previous approaches lies in our layer-wise embedding computation and the way similarity is computed between query and support images. As shown in (a), Previous work typically uses CNNs or Transformers to generate single-layer image embeddings; however, a single-layer embedding may not effectively integrate complex semantic information. This feature is then used to compute a single-layer correlation map, which is applied to reweight the image features. In contrast, as shown in (b), our method integrates multi-layer outputs from the backbone to form multi-level correlation maps. We then compute layer-wise weights for the features at different levels, creating an image embedding that captures diverse semantic focuses across different levels while avoiding the complexity of CNNs and Transformers. In (c), prior methods typically calculate the similarity between corresponding pixels at the same locations in both images. However, this approach overlooks the possibility that semantically similar pixels may be located in different positions, making it difficult to assess the true similarity between the image pairs accurately. In contrast, (d) our method employs a matching algorithm that identifies the most similar pixel in the support image for each pixel in the query image, even if their positions do not align perfectly. This allows for a more accurate evaluation of the true similarity score.}
\label{fig1}
\end{figure}

However, through our study of prior methods, we identified two key limitations in metric-based methods: (1) The different semantic focuses produced by various levels of the backbone have not been fully utilized. Features at different levels typically focus on distinct aspects of the image~\cite{a30}. In~\cite{a29,a20}, only features from the final layer of the backbone are utilized. However, this layer may be overly focused on a specific aspect of the image. In few-shot learning, it is crucial to leverage a wider variety of features to assess the similarity between image pairs more effectively. Chen et al. proposed using a self-attention mechanism to learn the relationships between multi-level features from the backbone but introduced excessive parameters and computational complexity~\cite{a21}. (2) Semantic pixel misalignment is also a common challenge. In these methods, the final query embedding and support embedding are typically compared using element-wise cosine similarity. However, directly performing element-wise comparisons can often fail to correctly match semantically similar pixels, as illustrated in Fig. ~\ref{fig1} (c).

To address the issue of utilizing semantic focus, we propose a Layer-wise Embedding (LWE) Module. By computing the correlation map generated from different levels of information and performing layer-wise production, we effectively integrate diverse semantic features without relying on convolution or self-attention operations. As a result, our approach reduces both the number of parameters and the computational complexity. However, after aggregating multi-level semantic features, we observed a persistent issue of positional inconsistency between semantically corresponding objects in paired images. To tackle this challenge, we rearrange the semantic pixel positions in the image pair by leveraging the Hungarian matching algorithm and a learnable matcher. This ensures that the most similar semantic pixels are aligned spatially, enabling the calculation of a more accurate similarity score, which further aids in classification. Our proposed method integrates these two modules by using the LWE to learn semantic embeddings from different levels and the SPM to achieve pixel-wise similarity matching, which creates an end-to-end process that takes an image pair as input and outputs a similarity score. 

To sum up, our main contributions are as follows:
\begin{itemize}\item[$\bullet$]We propose a novel Layer-wise Embedding Module that efficiently integrates semantic information from different levels with lightweight operations.\end{itemize}
\begin{itemize}\item[$\bullet$]We introduce the Semantic-Pixel Matching Module, which leverages the Hungarian algorithm and a learnable matcher to capture pixel-level correlations between image pairs, addressing the limitations of prior work in accurately assessing true similarity.\end{itemize}
\begin{itemize}\item[$\bullet$]Our proposed LWEM-SPM combines these two modules, and extensive experiments demonstrate that our method achieves competitive performance.\end{itemize}

%% The Appendices part is started with the command \appendix;
%% appendix sections are then done as normal sections
\section{Related works}
\subsection{Few-shot classification}
Few-shot classification methods have evolved significantly over the years and can be broadly categorized into four types. The first type is data augmentation-based methods~\cite{a1,a2,a3,a4,a5}, where the primary idea is to increase the number of samples for underrepresented classes, thereby helping the model achieve more accurate classification. However, our approach eliminates the need for complex data augmentation, making it simpler and more efficient. The second category comprises parameter optimization-based methods. In few-shot tasks, there is often a distribution mismatch between the training and test sets, causing the model to perform poorly on unseen data. To address this, parameter optimization-based methods~\cite{a6,a7,a8, a9, a10} often  a meta-learning strategy to optimize model parameters across different tasks. During training, the dataset is split into multiple tasks, enabling the model to better adapt to new tasks. The third type is transfer learning-based methods~\cite{a49,a42,a47}, which primarily focus on extracting knowledge from source domains and applying that knowledge to a target domain. The final category consists of metric-based methods~\cite{a11,a12,a13,a14,a15, b3,b4}, where the model learns to measure the similarity between different categories. It generates similar embeddings for the same class and more distant embeddings for various classes. At inference time, the distance between the query and support embeddings is used to classify the query. Our method also incorporates this concept, thus it can be classified as a metric-based approach.

\subsection{Cross-correlation}
Cross-correlation is a widely used technique for identifying relationships between two signals, such as images, making it valuable in various matching tasks~\cite{a16,a17}. In the context of few-shot learning, accurately determining the correlation between images has become increasingly important for improving model performance. As a result, cross-correlation has been adapted for few-shot tasks, finding applications in areas such as segmentation~\cite{a18,a19} and classification~\cite{a20,a21,a29}. 

However, previous methods have rarely explored multi-level cross-correlation, which can provide a more comprehensive focus on multi-level semantic information. Meanwhile, many of these methods compute cross-correlation by focusing solely on pixel similarity at corresponding positions, thereby neglecting the potential similarity between pixels located at different positions. 

In this paper, we address these issues by designing a multi-layer semantic aggregation method that integrates multi-level semantic focus with minimal additional complexity, while also exploring the application of assignment algorithms in computing pixel relationships.

\subsection{Assignment algorithm}
Assignment algorithms play a crucial role in various applications, enabling the optimal pairing of elements across domains. These algorithms have found significant utility in visual tasks in recent years~\cite{a22,a23,a24}, where accurate correspondences between images or features are essential for performance. Two commonly employed matching algorithms are the Hungarian Algorithm~\cite{a25} and the K-Nearest Neighbors (KNN)~\cite{a26} algorithm. The Hungarian Algorithm is well-known for solving the assignment problem by efficiently finding the optimal one-to-one matching between two sets. It minimizes the total cost of the assignments, making it particularly useful in resource allocation and task assignment in visual applications. On the other hand, the KNN algorithm is a versatile non-parametric method used for both classification and regression. By identifying the closest data points based on a defined distance metric, KNN facilitates the prediction of labels or values through the majority voting of neighbors or averaging of outcomes. Its simplicity and effectiveness have made it a popular choice in various visual tasks, particularly in low-dimensional settings.

In this paper, we conduct experiments on both the Hungarian Algorithm and KNN, leveraging their strengths to enhance the similarity assessment in our proposed method. A detailed analysis can be found in the experiment section.
\begin{figure*}[t]
\centerline{\includegraphics[width=\linewidth]{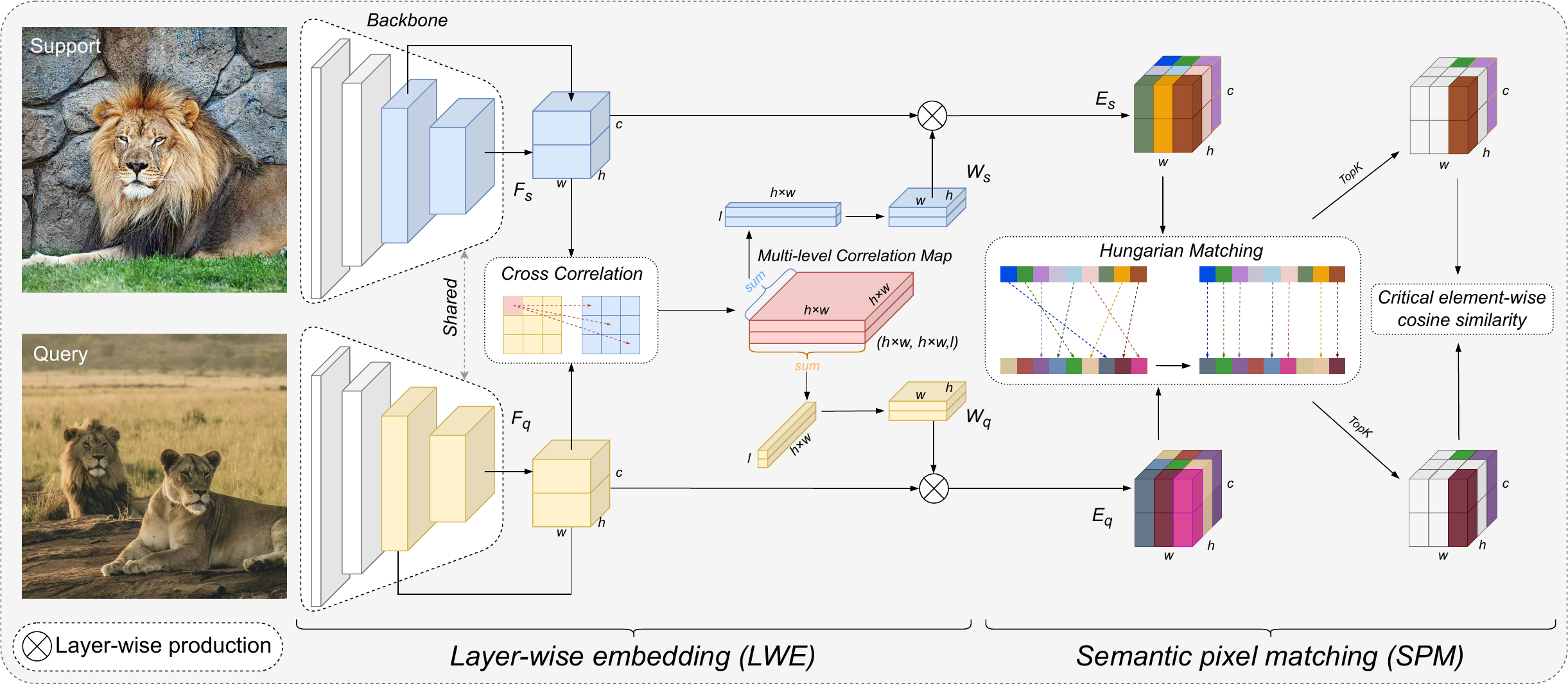 }}
\caption{Overview of proposed LWFM-SPM. The method consists of two stages, following a coarse-to-fine approach. First, Layer-wise Embedding (LWE) is used to generate multi-level correlation maps, producing well-focused semantic maps from the image pair. The Semantic-Pixel Matching (SPM) then provides fine-grained metrics for classification by reassigning semantic pixels between the feature maps of image pairs at each layer, using an assignment algorithm.
}
\label{fig2}
\end{figure*}
\section{Methodology}
In this section, we provide a detailed description of our proposed method. Fig. \ref{fig2} illustrates the overall architecture of our proposed Layered Weighted Feature Metric with Semantic Pixel Matching (LWFM-SPM). LWFM-SPM consists of two main modules: the layer-wise embedding Module and the semantic-pixel matching module. The first module generates well-focused feature maps for image pairs by producing semantic feature maps and cross-correlation matrices for support and query images at each layer and using the scores from the cross-correlation matrices to adjust the weight of each semantic feature in the feature maps. The second module provides fine-grained metrics for classification by reassigning semantic pixels between feature maps of image pairs using an assignment algorithm at each layer and then calculating pixel-level similarities.

\subsection{Architecture overview}
Given a support image \( I_{s} \) and a query image \( I_{q} \), we first use a pre-trained feature extractor to extract feature maps at different layers. Let \( E_{s} = \{ E^{l}_{s} \in \mathbb{R}^{h^{l} \times w^{l} \times c^{l}} \}_{l=1}^{L} \) and \( E_{q} = \{ E^{l}_{q} \in \mathbb{R}^{h^{l} \times w^{l} \times c^{l}} \}_{l=1}^{L} \) denote the sets of feature maps for the support and query images respectively. Here, \( h^{l} \times w^{l} \times c^{l} \) represents the shape of the feature map at the \( l \)-th layer of the feature extractor. Then we calculate the layer-wise cross-correlation matrix $corr = \{corr^{l} \in 
\mathbb{R}^{hw \times hw}\}^{L}_{l=1}$ between $E_{s}$ and $E_{q}$:
\begin{equation}
corr^{l}=reshape(pooling(E^{l}_{s})) \cdot reshape(pooling(E^{l}_{q}))^{T}
\end{equation}
where $\cdot$ indicates dot product, pooling($\cdot$) operation uses adaptive average pooling to map all feature maps to the shape of \( h \times w \times c^{l} \) and reshape($\cdot$) operation changes $ E^{l} \in \mathbb{R}^{h \times w \times c^{l}}$ to $\mathbb{R}^{hw \times c^{l}}$.  The correlation map $corr^{l}$ contains correlated information between the feature maps $E^{l}_{s}$ and $E^{l}_{q}$ and can be used to adjust the weights of critical features in those feature maps.  After generating the cross-correlation matrix $corr^{l}$, we need to compute the weight vectors $W^{l}_{s} \in \mathbb{R}^{wh \times 1}$ and $W^{l}_{q} \in \mathbb{R}^{wh \times 1}$ differently for the support and query images. This requires that the softmax function is applied in the correct direction:
\begin{equation}
W^{l}_{t}=\left\{
\begin{aligned}
W^{l}_{s,i}=\frac{\sum^{wh-1}_{0}e^{\nicefrac{corr^{l}_{i,k}}{\mathcal{T}}}}{\sum^{wh-1}_{0}e^{\nicefrac{corr^{l}_{k,j}}{\mathcal{T}}}} && \forall i,j \in [0,wh), t=s\\
W^{l}_{q,j}=\frac{\sum^{wh-1}_{0}e^{\nicefrac{corr^{l}_{k,j}}{\mathcal{T}}}}{\sum^{wh-1}_{0}e^{\nicefrac{corr^{l}_{i,k}}{\mathcal{T}}}} && \forall i,j \in [0,wh), t=q 
\end{aligned}
\right.
\end{equation}
where \(\mathcal{T}\) is a hyperparameter that regulates the sharpness of the softmax function, affecting the distribution of attention across the feature maps. The weight vectors reflect the importance of pixels at corresponding positions, with higher weights indicating greater significance of the features in the task context. We use these weight vectors to adjust the feature maps of the image:
\begin{gather}
{E^{l}_{s}}^{\prime}=reshape(W^{l}_{s}) \otimes E^{l}_{s} \\
{E^{l}_{q}}^{\prime}=reshape(W^{l}_{q}) \otimes E^{l}_{q}
\end{gather}
where $\otimes$ denotes Hadamard product, and reshape($\cdot$) operation changes $ W^{l} \in \mathbb{R}^{hw \times 1}$ to $\mathbb{R}^{h \times w \times 1}$.

${E^{l}_{s}}^{\prime}$ and ${E^{l}_{q}}^{\prime}$ are then fed into the semantic-pixel matching module to get the well-matched feature maps of the image pairs, ${E^{l}_{s}}^{\prime\prime}$ and ${E^{l}_{q}}^{\prime\prime}$.  Since not all features are critical features, we choose the top k pairs of features that match best as the metrics for our method using the following formula:
\begin{equation}
Score_{critical}=\sum^{k-1}_{i=0}\frac{{E^{l}_{s,i}}^{\prime\prime} \cdot {E^{l}_{q,i}}^{\prime\prime}}{\Vert{E^{l}_{s,i}}^{\prime\prime}\Vert\Vert{E^{l}_{q,i}}^{\prime\prime}\Vert}
\end{equation}
where \( i \) represents the feature pair with the \( i \)-th degree of matching.  

In order to preserve global valid information, we use the following formulas to calculate global scores between feature maps:
\begin{gather}
\Bar{E^{l}}^{\prime}=\frac{\sum^{h-1}_{i=0}\sum^{w-1}_{j=0}{E^{l}_{i,j}}^{\prime}}{hw}\\
Score_{global}=\frac{\Bar{{E^{l}_{s}}}^{\prime} \cdot \Bar{{E^{l}_{q}}}^{\prime}}{\Vert\Bar{{E^{l}_{s}}}^{\prime}\Vert\Vert\Bar{{E^{l}_{q}}}^{\prime}\Vert}
\end{gather}
Therefore, the final matching score between the two images is:
\begin{equation}
Score=\alpha \cdot Score_{critical} + Score_{global}
\end{equation}
where $\alpha$ is a hyperparameter to balance the contributions of critical feature matching and global information. By adjusting the value of $\alpha$, we can control the emphasis on critical features versus the overall context.  Since each layer of feature extraction may focus differently, we use the maximum value to process $Score_{critical}$ of different layers and the average value to process $Score_{global}$.

\subsection{Layer-wise embedding module}
Images should be embedded in a metric space before calculating the correlation score between image pairs as most metric-based methods do. Unlike previous methods that rely solely on the output of the final layer or merge multiple layers into one, our method isolates the feature maps of image pairs individually at each layer.  Our approach utilizes ResNet18~\cite{a28} as our feature backbone. ResNet18 consists of an initial convolutional layer and 8 basic blocks, each producing a feature map of different dimensionalities, represented as \( E^{l} \in \mathbb{R}^{h^{l} \times w^{l} \times c^{l}} \), where \( l \in \{0, 1, \dots, 8\} \) denotes the number of layers. We pre-train our feature backbone to accelerate the training process.
\begin{figure*}[t]
\centerline{\includegraphics[width=\linewidth]{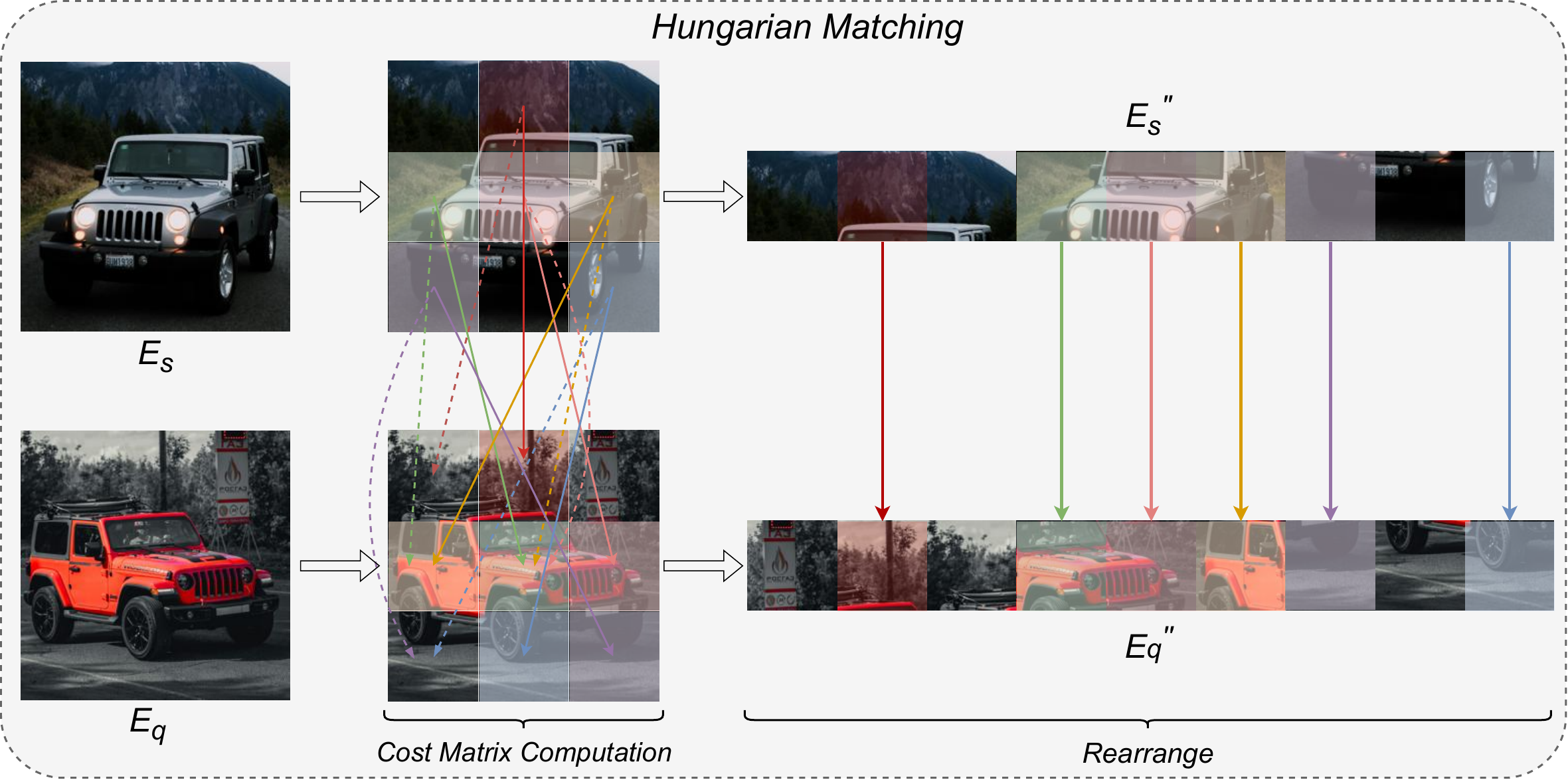}}
\caption{The overview of our proposed Hungarian matching algorithm consists of two stages: calculating the cost matrix to find the best matching pixels and rearranging the pixels. Given the input \(E_s\) and \(E_q\), we first compute a similarity matrix (i.e., cost matrix) between every pair of pixels. Using the Hungarian algorithm, we identify the combination that maximizes global similarity. Then, \(E_q\) is rearranged according to the matching results, ensuring that similar pixels are aligned. This allows us to accurately compute the true similarity between the two images in subsequent steps. 
}
\label{fig3}
\end{figure*}
\subsection{Semantic-pixel matching module}
In this section, we describe the details of our semantic-pixel matching module.  In the semantic-pixel matching module, our goal is to perform fine-grained matching based on the adjusted feature maps ${E^{l}_{s}}^{\prime}$ and ${E^{l}_{q}}^{\prime}$. We employ an optimal assignment algorithm to reassign pixels from the query image to the pixels of the support image in order to form the best pairing. Specifically, we construct a spatial mapping by defining a matching matrix \( M^l \) to represent the matching degree of each pixel pair. The matching degree \( M_{i,j}^l \in \mathbb{R}^{hw \times hw}\) can be expressed as:
\begin{gather}
{E^{l}}^{\prime}=reshape({E^{l}}^{\prime})\\
M^{l}_{i,j}=\frac{{E^{l}_{s,i}}^{\prime} \cdot {E^{l}_{q,j}}^{\prime}}{\Vert{E^{l}_{s,i}}^{\prime}\Vert\Vert{E^{l}_{q,j}}^{\prime}\Vert}
\end{gather}
where the reshape($\cdot$) operation changes $ {E^{l}}^{\prime} \in \mathbb{R}^{h \times w \times c^l}$ to $\mathbb{R}^{hw \times c^l}$. 

We utilize the Hungarian algorithm to find the optimal assignment between feature pairs from the support and query feature maps. The following algorithm describes the detailed steps to achieve this:
\begin{algorithm}[H]
\caption{Optimal Feature Matching via Hungarian Algorithm}
\label{alg:hungarian}
\begin{algorithmic}[1]
    \STATE \textbf{Input:} Matched degree matrix $M^{l} \in \mathbb{R}^{hw \times hw}$
    \STATE \textbf{Output:} Indices of matched feature pairs $(i, j)$
    
    \STATE Compute the cost matrix $C^{l} = 1 - M^{l}$ \COMMENT{Convert matching degree to cost}
    
    \STATE Initialize assignment vector $A \in \mathbb{R}^{hw}$
    
    \STATE \textbf{Apply the Assignment algorithm:}
    \STATE \textbf{Step 1:} Transform the cost matrix $C^{l}$ if necessary (e.g., subtracting row minima)
    
    \STATE \textbf{Step 2:} Select the minimum cost assignment using the Hungarian algorithm:
    \STATE \textbf{Let:} $A(i)$ denote the matched index for feature $i$ 
    \FOR{each row $i$ in $C^{l}$}
        \STATE Find the column index $j$ such that $C^{l}_{i,j}$ is minimized
        \STATE Assign $A(i) \gets j$
    \ENDFOR
    
    \STATE \textbf{Step 3:} Return indices $(i, A(i))$ for all $i$
    
    \STATE \textbf{Output:} Matched indices $(i, A(i))$ representing optimal pixel matches between support and query feature maps

\end{algorithmic}
\end{algorithm}

Then we rearrange ${E^{l}_{s}}^{\prime}$ and ${E^{l}_{q}}^{\prime}$ according to the index to get ${E^{l}_{s}}^{\prime\prime}$ and ${E^{l}_{q}}^{\prime\prime}$. 

To fully capture the complex relationships between pixels across feature maps, we introduce a learnable matcher designed as a bottleneck MLP structure, which can be expressed as follows:
\begin{gather}
{E^{l}_{s}}^{\prime\prime} = \sigma_2(Linear(\sigma_1(Linear({E^{l}_{s}}^{\prime\prime})))) + {E^{l}_{s}}^{\prime\prime}\\
{E^{l}_{q}}^{\prime\prime} = \sigma_2(Linear(\sigma_1(Linear({E^{l}_{q}}^{\prime\prime})))) + {E^{l}_{q}}^{\prime\prime}
\end{gather}
where \(\sigma\) represents the ReLU activation function, with \(Linear_1\) as a linear layer that halves the number of channels, and \(Linear_2\) restores the channel dimensions.

\subsection{Learning objective}
In our method, the score is the basis for the metric-based classifier's judgment, so we can get the metric-based loss $\mathcal{L}_1$:
\begin{equation}
\mathcal{L}_1=-log\frac{exp(Score(E_s^{(n)},E_q^{(n)}))}{\sum^N_{n^{\prime}}exp(Score(E_s^{(n^{\prime})},E_q^{(n^{\prime})}))}
\end{equation}
where $Score(\cdot,\cdot)$ is the function that gets the final matching score from the feature maps set of the image pair. Notably, a query image has $N \times K$ different embeddings because there are $N \times K$ support images in one task, We average $K$ final matching score. 

Following RENet~\cite{a20} and CAN~\cite{a29}, we define an enhanced classifier to enhance the classification accuracy. We refer to the loss produced by the enhanced classifier as $\mathcal{L}_2$:
\begin{equation}
\mathcal{L}_2=-log\frac{exp(w^{T}\Bar{E}^{\prime}_q+b)^{(c)}}{\sum^{|\mathcal{C}_{train}|}_{c^{\prime}=1}exp(w^{T}\Bar{E}^{\prime}_q+b)^{(c^{\prime})}}
\end{equation}
where $w$ and $b$ are the weight and bias of a fully-connected layer.  Therefore, the learning objective 
$\mathcal{L}$ combines two components, where $\beta$ is a hyper-parameter to balance these two components:
\begin{equation}
\mathcal{L}=\beta \cdot \mathcal{L}_1 + \mathcal{L}_2
\end{equation}

\section{Experiments}
\subsection{Datasets}
We utilize the four most commonly used datasets in few-shot learning: miniImageNet, tieredImageNet, CUB-200-2011, and CIFAR-FS. Vinyals et al. created the miniImageNet~\cite{a31} dataset by extracting a subset of images from the ImageNet~\cite{a36} dataset, consisting of 100 classes with 600 images per class. The dataset is split into 64 classes for training, 16 for validation, and 20 for testing. tieredImageNet~\cite{a32} includes 608 classes and 779,165 images from ImageNet, with train/validation/test splits of 351/97/160 classes, respectively. The CUB-200-2011~\cite{a33} dataset is widely used for fine-grained visual categorization tasks, containing 11,788 images across 200 bird subcategories, with splits of 100/50/50 classes for training, validation, and testing. CIFAR-FS~\cite{a34} is based on CIFAR-100~\cite{a35}, which contains 60,000 images and comprises 64/16/20 classes for training, validation, and testing.
\subsection{Implemention details}
We adopted a ResNet18 backbone pre-trained on ImageNet. For LWE, we empirically selected the last two layers' outputs of backbone, ie. \( l \in \{7, 8\} \). Both outputs are resized to 3$\times$3 using adaptive pooling. 

In \(\mathcal{L}\), the hyperparameter \(\beta\) was determined through grid search, resulting in values of 0.25, 0.25, 0.5, and 1.5 for Mini-ImageNet, TieredImageNet, CIFAR-FC, and CUB-200-2011, respectively. Regarding \(\alpha\) in \(Score\), we considered that the backbone is pre-trained on ImageNet, with Mini-ImageNet and TieredImageNet being subsets. Introducing too many parameters may disrupt the inherent inductive bias of the backbone; therefore, we set \(\alpha\) to 0.25 for these two datasets, and for CIFAR-FC and CUB-200-2011, \(\alpha\) is set to 1. The temperature factor \(\mathcal{T}\) is uniformly set to 5. For the N-way K-shot evaluation, we test 15 query samples for each class in an episode and report the average classification accuracy with 95\% confidence intervals over 2,000 randomly sampled test episodes. Our model is trained on an NVIDIA RTX 4070 Ti for 30 epochs on the CUB-200-2011 dataset and 10 epochs on the other datasets, with an initial learning rate of 0.01. The learning rate decays by a factor of 0.05 at epochs 20, 24, 26, and 28 for CUB-200-2011, and at epochs 4, 6, and 8 for the others.

\subsection{Comparison Methods}
We conducted comparisons between LWFM-SPM and a diverse range of approaches, including Data augmentation-based~\cite{a60}, Parameter optimization-based~\cite{a59, a52, a49, a43}, Metric-based~\cite{a40,a41,a46,a44,a48,a29,a50,a54,a53,a58,a57,a51,a20,a21}, and
Other methods~\cite{a49,a42,a47}. Frequently
used for benchmarking, these approaches have
consistently demonstrated enduring performance
over time.

Among these methods, MCNet~\cite{a21} represents the best performance in metric-based approaches. However, its use of self-attention mechanism results in an excessive number of parameters and increased computational complexity.

\subsection{Qualitative analysis}
Tables \ref{t1},\ref{t2},\ref{t3} and \ref{t4} compare LWFM-SPM and various few-shot classification methods on four datasets. Notably, our method uses the same or smaller backbone compared to methods~\cite{a60,a46,a47,a50,a21,a58,a59,a49,a57}, yet outperforms them, especially on the CIFAR-FS and CUB-200-2011 datasets. Among the methods that use the same backbone as ours~\cite{a60,a46,a47,a50,a21}, we focus on comparing with the previous state-of-the-art, MCNet, which also explored the cross-correlation mechanism. Our method outperforms MCNet in 5-way 1-shot test accuracy on the miniImageNet, TieredImageNet, CUB-200-2011, and CIFAR-FS datasets by 0.10, 0.10, 0.35, and 0.68, respectively. Notably, our advantage on CIFAR-FS and CUB-200-2011 is greater than that on miniImageNet and tieredImageNet. This can be attributed to the inductive bias of the pre-trained ResNet18, which is already well-tuned for images from the ImageNet subset. Thus, models trained on miniImageNet and TieredImageNet benefit less from additional improvements, while the diversity of CIFAR-FS and CUB-200-2011 enables our method to achieve more substantial gains. 
\begin{table}[t!]
\renewcommand\arraystretch{1.15}
\begin{center}
\caption{Performance comparison in terms of accuracy (\%) on miniImageNet.}
\resizebox{0.5\linewidth}{!}{
\begin{tabular}{cccc}
\toprule
Methods      & Backbone  & 5-way 1-shot     & 1-way 1-shot \\ \midrule
Shot-Free~\cite{a40}    & ResNet12  & 59.04            & 77.64 \\
TPN~\cite{a41}          & ResNet12  & 59.46            & 75.65 \\
MTL~\cite{a45}          & ResNet12  & 61.20 $\pm$ 1.80 & 75.50 $\pm$ 0.80 \\
RFS-simple~\cite{a42}   & ResNet12  & 62.02 $\pm$ 0.63 & 79.64 $\pm$ 0.44 \\
AFHN~\cite{a60}         & ResNet18  & 62.38 $\pm$ 0.72 & 78.16 $\pm$ 0.56 \\
DC~\cite{a46}           & ResNet18  & 62.53 $\pm$ 0.19 & 79.77 $\pm$ 0.19 \\
ProtoNet~\cite{a44}     & ResNet12  & 62.39 $\pm$ 0.21 & 80.53 $\pm$ 0.14 \\
MetaOptNet~\cite{a43}   & ResNet12  & 62.64 $\pm$ 0.82 & 78.63 $\pm$ 0.46 \\ 
SimpleShot~\cite{a47}   & ResNet18  & 62.85 $\pm$ 0.20 & 80.02 $\pm$ 0.14 \\
MatchNet~\cite{a48}     & ResNet12  & 63.08 $\pm$ 0.80 & 75.99 $\pm$ 0.60 \\
CAN~\cite{a29}          & ResNet12  & 63.85 $\pm$ 0.48 & 79.44 $\pm$ 0.34 \\ 
NegMargin~\cite{a49}    & ResNet12  & 63.85 $\pm$ 0.81 & 81.57 $\pm$ 0.56 \\
CTM~\cite{a50}          & ResNet18  & 64.12 $\pm$ 0.82 & 80.51 $\pm$ 0.13 \\
MLCN~\cite{a51}         & ResNet12  & 65.54 $\pm$ 0.43 & 81.63 $\pm$ 0.31 \\
DeepEMD~\cite{a54}      & ResNet12  & 65.91 $\pm$ 0.82 & 82.41 $\pm$ 0.56 \\
FEAT~\cite{a53}         & ResNet12  & 66.78 $\pm$ 0.20 & 82.05 $\pm$ 0.14 \\
GLFA~\cite{a52}         & ResNet12  & 67.25 $\pm$ 0.36 & 82.80 $\pm$ 0.30 \\
MCL-Katz~\cite{a55}     & ResNet12  & 67.51            & 83.99 \\
CVET~\cite{a56}         & ResNet12  & 70.19 $\pm$ 0.46 & 84.66 $\pm$ 0.29 \\
RENet~\cite{a20}        & ResNet12  & 67.60 $\pm$ 0.44 & 82.58 $\pm$ 0.30 \\
MCNet~\cite{a21}        & ResNet18  & \underline{72.13 $\pm$ 0.43} & \underline{88.20 $\pm$ 0.23} \\
LWFM-SPM(Ours)          & ResNet18  & \textbf{72.22 $\pm$ 0.43} & \textbf{88.22 $\pm$ 0.24} \\
\bottomrule
\end{tabular}
}
\label{t1}
\end{center}
\end{table}
\begin{table}[t!]
\renewcommand\arraystretch{1.15}
\begin{center}
\caption{Performance comparison in terms of accuracy (\%) on tieredImageNet.}
\resizebox{0.5\linewidth}{!}{
\begin{tabular}{cccc}
\toprule
Methods      & Backbone  & 5-way 1-shot     & 1-way 1-shot \\ \midrule
Shot-Free~\cite{a40}    & ResNet12  & 63.52            & 82.59\\
TPN~\cite{a41}          & ResNet12  & 59.91 $\pm$ 0.94 & 73.30 $\pm$ 0.75 \\
RFS-simple~\cite{a42}   & ResNet12  & 69.74 $\pm$ 0.72 & 84.41 $\pm$ 0.55 \\
ProtoNet~\cite{a44}     & ResNet12  & 68.23 $\pm$ 0.23 & 84.03 $\pm$ 0.16 \\
MetaOptNet~\cite{a43}   & ResNet12  & 65.99 $\pm$ 0.72 & 81.56 $\pm$ 0.53 \\ 
MatchNet~\cite{a48}     & ResNet12  & 68.50 $\pm$ 0.92 & 80.60 $\pm$ 0.71 \\ 
CAN~\cite{a29}          & ResNet12  & 69.89 $\pm$ 0.51 & 84.23 $\pm$ 0.37 \\ 
CTM~\cite{a50}          & ResNet12  & 68.41 $\pm$ 0.39 & 84.28 $\pm$ 1.73 \\ 
MLCN~\cite{a51}         & ResNet12  & 71.62 $\pm$ 0.49 & 85.58 $\pm$ 0.35 \\
DeepEMD~\cite{a54}      & ResNet12  & 71.16 $\pm$ 0.87 & 86.03 $\pm$ 0.58 \\
FEAT~\cite{a53}         & ResNet12  & 70.80 $\pm$ 0.23 & 84.79 $\pm$ 0.16 \\
GLFA~\cite{a52}         & ResNet12  & 72.25 $\pm$ 0.40 & 86.37 $\pm$ 0.27 \\
MCL-Katz~\cite{a55}     & ResNet12  & 72.01            & 86.02 \\
CVET~\cite{a56}         & ResNet12  & 72.62 $\pm$ 0.51 & 86.62 $\pm$ 0.33 \\ 
RENet~\cite{a20}        & ResNet12  & 71.61 $\pm$ 0.51 & 85.28 $\pm$ 0.35 \\
MCNet~\cite{a21}        & ResNet18  & \underline{73.14 $\pm$ 0.46} & \underline{87.16 $\pm$ 0.32} \\
LWFM-SPM(Ours)          & ResNet18  & \textbf{73.24 $\pm$ 0.47} & \textbf{87.38 $\pm$ 0.31} \\
\bottomrule
\end{tabular}
}
\label{t2}
\end{center}
\end{table}
\begin{table}[t!]
\renewcommand\arraystretch{1.15}
\begin{center}
\caption{Performance comparison in terms of accuracy (\%) on CUB-200-2011.}
\resizebox{0.5\linewidth}{!}{
\begin{tabular}{cccc}
\toprule
Methods      & Backbone  & 5-way 1-shot     & 1-way 1-shot \\ \midrule
RelationNet~\cite{a58}  & ResNet34  & 66.20 $\pm$ 0.99 & 82.30 $\pm$ 0.58 \\
MAML~\cite{a59}         & ResNet34  & 67.28 $\pm$ 1.08 & 83.47 $\pm$ 0.59 \\
AFHN~\cite{a60}         & ResNet18  & 70.53 $\pm$ 1.01 & 83.95 $\pm$ 0.63 \\
NegMargin~\cite{a49}    & ResNet18  & 72.66 $\pm$ 0.85 & 89.40 $\pm$ 0.43 \\
MatchNet~\cite{a47}     & ResNet18  & 71.87 $\pm$ 0.85 & 85.08 $\pm$ 0.57 \\
DeepEMD~\cite{a54}      & ResNet12  & 75.65 $\pm$ 0.83 & 88.69 $\pm$ 0.50 \\
FEAT~\cite{a53}         & ResNet12  & 73.27 $\pm$ 0.22 & 85.77 $\pm$ 0.14 \\
ProtoNet~\cite{a44}     & ResNet12  & 66.09 $\pm$ 0.92 & 82.50 $\pm$ 0.58 \\
S2M2~\cite{a57}         & ResNet34  & 72.92 $\pm$ 0.83 & 86.55 $\pm$ 0.51 \\ 
GLFA~\cite{a52}         & ResNet12  & 76.52 $\pm$ 0.37 & 90.27 $\pm$ 0.38 \\
MLCN~\cite{a51}         & ResNet12  & 77.96 $\pm$ 0.44 & 91.20 $\pm$ 0.24 \\
RENet~\cite{a20}        & ResNet12  & 79.49 $\pm$ 0.44 & 91.11 $\pm$ 0.24 \\
MCNet~\cite{a21}        & ResNet18  & \underline{81.72 $\pm$ 0.43} & \underline{92.62 $\pm$ 0.23} \\
LWFM-SPM(Ours)          & ResNet18  & \textbf{82.07 $\pm$ 0.43} & \textbf{92.87 $\pm$ 0.23} \\
\bottomrule
\end{tabular}
}
\label{t3}
\end{center}
\end{table}
\begin{table}[t!]
\renewcommand\arraystretch{1.15}
\begin{center}
\caption{Performance comparison in terms of accuracy (\%) on CIFAR-FS.}
\resizebox{0.5\linewidth}{!}{
\begin{tabular}{cccc}
\toprule
Methods      & Backbone  & 5-way 1-shot     & 1-way 1-shot \\ \midrule
Shot-Free~\cite{a40}    & ResNet12  & 69.2             & 84.7 \\
RFS-simple~\cite{a42}   & ResNet12  & 71.5 $\pm$ 0.8   & 86.0 $\pm$ 0.5 \\
MetaOptNet~\cite{a43}   & ResNet12  & 72.6 $\pm$ 0.7   & 84.3 $\pm$ 0.5 \\ 
ProtoNet~\cite{a44}     & ResNet12  & 72.2 $\pm$ 0.7   & 83.5 $\pm$ 0.5 \\
S2M2~\cite{a57}         & ResNet34  & 62.77 $\pm$ 0.23 & 75.75 $\pm$ 0.13 \\ 
CVET~\cite{a56}         & ResNet12  & 77.56 $\pm$ 0.47 & 88.64 $\pm$ 0.31 \\ 
GLFA~\cite{a52}         & ResNet12  & 74.01 $\pm$ 0.40 & 87.02 $\pm$ 0.27 \\
MLCN~\cite{a51}         & ResNet12  & 74.36 $\pm$ 0.46 & 87.24 $\pm$ 0.31 \\
RENet~\cite{a20}        & ResNet12  & 74.51 $\pm$ 0.46 & 86.60 $\pm$ 0.32 \\
MCNet~\cite{a21}        & ResNet18  & \underline{78.26 $\pm$ 0.44} & \underline{90.58 $\pm$ 0.29} \\
LWFM-SPM(Ours)          & ResNet18  & \textbf{78.94 $\pm$ 0.44} & \textbf{90.73 $\pm$ 0.30} \\
\bottomrule
\end{tabular}
}
\label{t4}
\end{center}
\end{table}
\begin{table}[t!]
\renewcommand\arraystretch{1.15}
\begin{center}
\caption{Comparison of parameter counts across several methods. Values in () represent the additional parameters excluding the backbone.}
\resizebox{0.8\linewidth}{!}{
\begin{tabular}{cccccccc}
\toprule
Methods      & CAN  & ProtoNet     & MatchNet  & FEAT & RENet & MCNet & LWFM-SPM(Ours)\\ \midrule
Parameters      & 12.02M(3.98M)  & 11.19M(0M)     & 11.19M(0M)  &14.06M(1.64M)  & 12.67M(0.24M) & 12.12M(1.0M) & 11.77M(0.65M)\\
\bottomrule
\end{tabular}
}
\label{t5}
\end{center}
\end{table}

Next, we analyze the computational complexity and time efficiency. MCNet utilizes a self-attention mechanism to fuse multi-layer correlation maps, an operation we find redundant and cumbersome. Compared to our layer-wise embedding approach, MCNet not only fails to offer a performance advantage but also incurs a much higher computational cost, as demonstrated in the following proof and params comparison in Table \ref{t5}. 

Similar to MCNet, in our method, the input images are transformed into \(d \times d \enspace (d=3)\) feature maps after passing through the backbone. The self-attention mechanism in MCNet operates on the \(d^2 \times d^2\) correlation map, resulting in a time complexity of:
\begin{equation}
O((d^2)^2 \cdot (d^2)^2) = O(d^8)
\end{equation}

In contrast, our method applies Hungarian matching between the two \(d \times d\) feature maps. During the cost matrix calculation, the time complexity is:
\begin{equation}
O((d^2)^2) = O(d^4)
\end{equation}
and in the matching stage, the complexity is:
\begin{equation}
O((d^2)^3) = O(d^6)
\end{equation}
Thus, the total complexity is:
\begin{equation}
O(d^4 + d^6) = O(d^6)
\end{equation}

It is evident that our method offers an advantage in terms of time complexity. For instance, during 5-way 1-shot training on the miniImageNet dataset, our method takes 24 seconds per epoch, whereas MCNet takes 48 seconds on the same NVIDIA RTX 4070 Ti GPU. On other datasets, we consistently maintain approximately twice the training speed of MCNet.
\subsection{Ablation studies}
To validate the effectiveness of our proposed layer-wise embedding and semantic pixel matching, we conducted extensive ablation studies. 

First, we experimented with the number of selected layers for our layer-wise embedding. The ResNet18 backbone we used has eight basic blocks, and we labeled their outputs as [1, 2, 3, 4, 5, 6, 7, 8]. We conducted experiments using seven different combinations of these blocks on the miniImageNet and CIFAR-FS datasets. As shown in Table \ref{t6}, we found that simply increasing the number of layers does not always lead to performance improvements. This is evident from the combination [1 2 3 4 5 6 7 8], which shows a slight improvement on miniImageNet but a significant drop on CIFAR-FS. Overall, the combination [ - - - - - - 7 8] proved to be the best choice, consistently demonstrating excellent performance across different datasets and being more robust than other combinations.
\begin{table}[h]
\setlength\tabcolsep{12pt} 
\renewcommand\arraystretch{1.15}
\begin{center}
\caption{Performance comparison in terms of 5-way 1-shot accuracy (\%) for the choices of layers in LWE.}
\resizebox{0.5\linewidth}{!}{
\begin{tabular}{ccc}
\toprule
Layers      & miniImageNet  & CIFAR-FS \\ \midrule
\texttt{[- - - - - - - 8]}   & 70.92 $\pm$ 0.426  & 78.32 $\pm$ 0.442  \\
\texttt{[- - - - - - 7 8]}   & \underline{72.22 $\pm$ 0.430}  & \textbf{78.94 $\pm$ 0.436} \\
\texttt{[- - - - - 6 7 8]}   & 71.87 $\pm$ 0.432  & 78.53 $\pm$ 0.440 \\ 
\texttt{[- - - - 5 6 7 8]}   & 72.17 $\pm$ 0.433  & 78.33 $\pm$ 0.439 \\
\texttt{[- 2 - 4 - 6 - 8]}   & 71.63 $\pm$ 0.427  & \underline{78.63 $\pm$ 0.442}\\ 
\texttt{[- - - 4 - 6 - 8]}   & 72.08 $\pm$ 0.430  & 78.49 $\pm$ 0.441\\ 
\texttt{[1 2 3 4 5 6 7 8]}   & \textbf{72.43 $\pm$ 0.430}  & 76.89 $\pm$ 0.444\\
\bottomrule
\end{tabular}
}
\label{t6}
\end{center}
\end{table}

Second, we experimented with the choice of assignment algorithm, comparing the Hungarian algorithm and the Nearest Neighbor algorithm, as shown in Table 7. It is evident that the Hungarian algorithm outperforms Nearest Neighbor. This is because the Hungarian algorithm considers global similarity, finding an optimal combination by minimizing the total global cost. In contrast, Nearest Neighbor operates on a local basis, selecting matches based on the nearest similarity at each point. While this local strategy can lead to quicker decisions, it often overlooks the broader context of the image pairs, leading to suboptimal global alignment. By considering the entire feature map, Hungarian matching ensures that the final assignment achieves the lowest possible cumulative cost, leading to more accurate similarity evaluations and improved overall performance.
\begin{table}[h]
\setlength\tabcolsep{3pt} 
\renewcommand\arraystretch{1.15}
\begin{center}
\caption{Performance comparison in terms of accuracy (\%) for the choices of assignment algorithm in SPM.}
\resizebox{0.5\linewidth}{!}{
\begin{tabular}{lcccc}
\toprule
\multirow{2}{*}{Method} & \multicolumn{2}{c}{miniImageNet} & \multicolumn{2}{c}{CIFAR-FS} \\ 
\cmidrule(lr){2-3} \cmidrule(lr){4-5}
& 5-way 1-shot & 5-way 5-shot & 5-way 1-shot & 5-way 5-shot  \\ \midrule
Hungarian   & \textbf{72.22 $\pm$ 0.43} & \textbf{88.22 $\pm$ 0.24} & \textbf{78.94 $\pm$ 0.44} & \textbf{90.73 $\pm$ 0.30} \\
Nearest-Neighbour   & 71.75 $\pm$ 0.44 & 87.92 $\pm$ 0.23 & 78.71 $\pm$ 0.44 & 90.00 $\pm$ 0.30\\
\bottomrule
\end{tabular}
}
\label{t7}
\end{center}
\end{table}
\begin{table*}[t!]
\renewcommand\arraystretch{1.15}
\begin{center}
\caption{Effect of Layer-Wise Embedding (LWE) and Semantic Pixel Matching (SPM) cross multiple datasets.}
\resizebox{\linewidth}{!}{
\begin{tabular}{lcccccccc}
\toprule
\multirow{2}{*}{Method} & \multicolumn{2}{c}{miniImageNet} & \multicolumn{2}{c}{tieredImagenet} & \multicolumn{2}{c}{CUB-200-2011} & \multicolumn{2}{c}{CIFAR-FS}\\ 
\cmidrule(lr){2-3} \cmidrule(lr){4-5} \cmidrule(lr){6-7} \cmidrule(lr){8-9}
                        & 5-way 1-shot & 5-way 5-shot & 5-way 1-shot & 5-way 5-shot & 5-way 1-shot & 5-way 5-shot & 5-way 1-shot & 5-way 5-shot\\ 
\midrule
ResNet18           & 70.25 $\pm$ 0.43  & 85.37 $\pm$ 0.25  & 70.284 $\pm$ 0.48 & 84.89 $\pm$ 0.33  & 80.19 $\pm$ 0.45  & 91.40 $\pm$ 0.24  & 77.05 $\pm$ 0.46  & 89.11 $\pm$ 0.30\\
LWE             & 72.20 $\pm$ 0.43  & 87.68 $\pm$ 0.23  & 73.11 $\pm$ 0.47  & 87.30 $\pm$ 0.31  & 81.69 $\pm$ 0.43  & 92.55 $\pm$ 0.23  & 78.56 $\pm$ 0.44  & 90.60 $\pm$ 0.29\\
SPM             & 71.79 $\pm$ 0.43  & 87.59 $\pm$ 0.25  & 72.93 $\pm$ 0.48  & 86.64 $\pm$ 0.32  & 81.29 $\pm$ 0.43  & 92.06 $\pm$ 0.23  & 78.51 $\pm$ 0.44  & 90.14 $\pm$ 0.30\\
LWE + SPM       & \textbf{72.22 $\pm$ 0.43}  & \textbf{88.22 $\pm$ 0.24}  & \textbf{73.24 $\pm$ 0.47}  & \textbf{87.38 $\pm$ 0.31}  &\textbf{82.07 $\pm$ 0.43}  & \textbf{92.87 $\pm$ 0.23}  & \textbf{78.94 $\pm$ 0.44}  & \textbf{90.73 $\pm$ 0.30}\\
\bottomrule
\end{tabular}
}
\label{t8}
\end{center}
\end{table*}
\begin{table*}[t!]
\renewcommand\arraystretch{1.15}
\begin{center}
\caption{Performance comparison with MCNet on different Backbones.}
\resizebox{\linewidth}{!}{
\begin{tabular}{lcccccccccc}
\toprule
\multirow{2}{*}{Method} &\multirow{2}{*}{Backbone} & \multicolumn{2}{c}{miniImageNet} & \multicolumn{2}{c}{tieredImageNet} & \multicolumn{2}{c}{CUB-200-2011} & \multicolumn{2}{c}{CIFAR-FS}\\ 
\cmidrule(lr){3-4} \cmidrule(lr){5-6} \cmidrule(lr){7-8} \cmidrule(lr){9-10}
 & & 5-way 1-shot & 5-way 5-shot & 5-way 1-shot & 5-way 5-shot & 5-way 1-shot & 5-way 5-shot & 5-way 1-shot & 5-way 5-shot\\ 
\midrule
ResNet34  &\_ &73.83 $\pm$ 0.429 &88.22 $\pm$ 0.240 & 72.75 $\pm$ 0.483 & 86.78 $\pm$ 0.305 & 81.79 $\pm$ 0.432  & 92.82 $\pm$ 0.230 & 76.11 $\pm$ 0.465  & 89.07 $\pm$ 0.293  \\
MCNet &ResNet34 & 74.35 $\pm$ 0.429 & 88.60 $\pm$ 0.291 & 72.65 $\pm$ 0.480 & 86.68 $\pm$ 0.308 & 82.26 $\pm$ 0.434 & \textbf{93.42} $\pm$ \textbf{0.224} & 77.21 $\pm$ 0.459 & 89.77 $\pm$ 0.306\\
LWFM-SPM(ours) &ResNet34 &\textbf{75.67} $\pm$ \textbf{0.421} & \textbf{89.87} $\pm$ \textbf{0.213} &\textbf{75.60} $\pm$ \textbf{0.465} &\textbf{88.98} $\pm$ \textbf{0.284}  & \textbf{83.43 }$\pm$ \textbf{0.409}  & 93.31 $\pm$ 0.224  &  \textbf{80.93} $\pm$ \textbf{0.424}  & \textbf{91.71} $\pm$ \textbf{0.284}  \\
\midrule
ResNet50  &\_ &80.88 $\pm$ 0.392 & 93.00 $\pm$ 0.160 &78.14 $\pm$ 0.453  & 90.24 $\pm$ 0.260  & 84.72 $\pm$ 0.401  & 94.18 $\pm$ 0.207  & 81.79 $\pm$ 0.432  & \textbf{92.82} $\pm$ \textbf{0.230}\\
MCNet &ResNet50 & 80.94 $\pm$ 0.394 & 93.04 $\pm$ 0.160 & 78.23 $\pm$ 0.451 & 90.46 $\pm$ 0.259 & 84.67 $\pm$ 0.409 & \textbf{94.40} $\pm$ \textbf{0.206} & 81.30 $\pm$ 0.443 & 92.64 $\pm$ 0.264\\
LWFM-SPM(ours) &ResNet50 &\textbf{81.42} $\pm$ \textbf{0.387} & \textbf{93.36} $\pm$ \textbf{0.165} &\textbf{79.26} $\pm$ \textbf{0.449} &\textbf{91.64} $\pm$ \textbf{0.248}  & \textbf{84.84} $\pm$ \textbf{0.393}  & 94.04 $\pm$ 0.212  &  \textbf{83.31} $\pm$ \textbf{0.420}  & 92.75 $\pm$ 0.264  \\
\bottomrule
\end{tabular}
}
\label{t9}
\end{center}
\end{table*}
\begin{figure*}[t]
 \centering
 \begin{subfigure}{0.32\linewidth}
     \includegraphics[width=\textwidth]{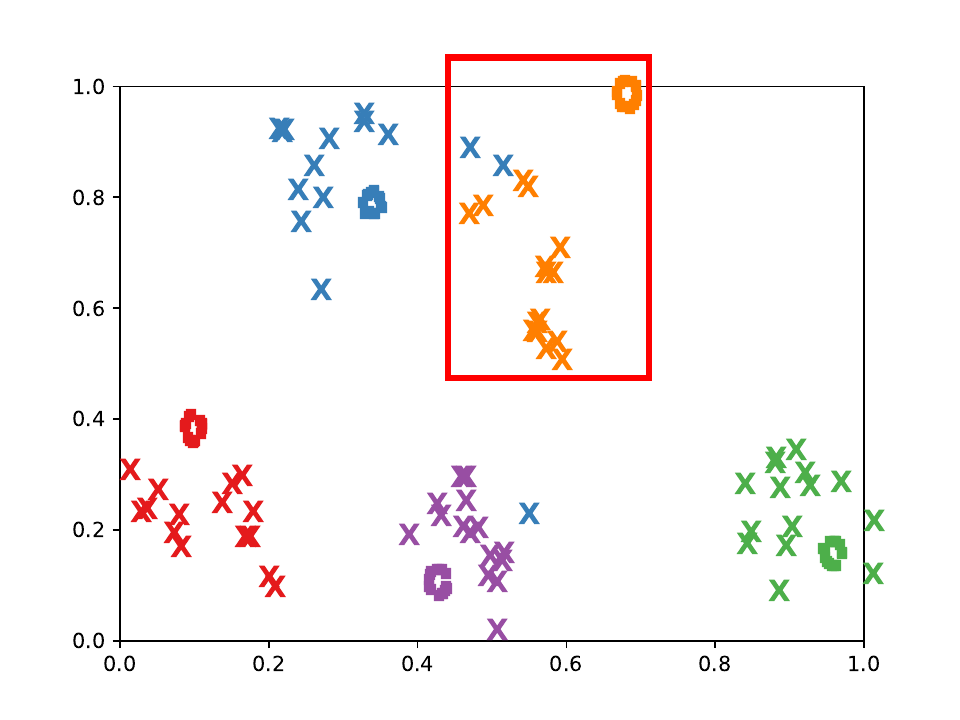}
     \caption{LWE on CIFAR-FS}
 \end{subfigure}
 \begin{subfigure}{0.32\linewidth}
     \includegraphics[width=\textwidth]{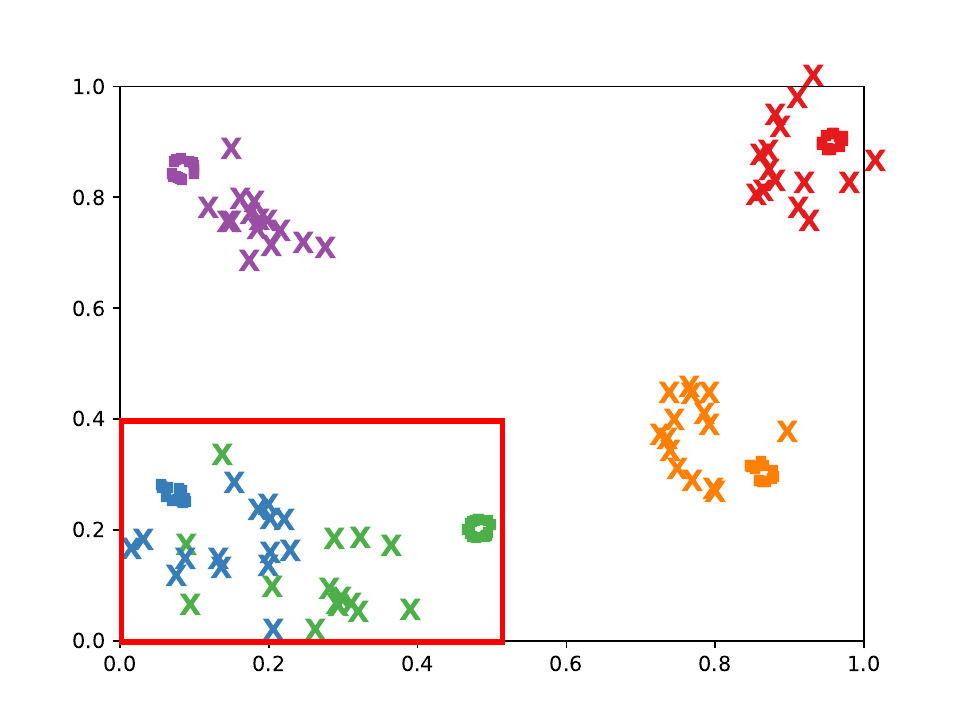}
     \caption{LWE on CUB-200-2011}
 \end{subfigure}
 \begin{subfigure}{0.32\linewidth}
     \includegraphics[width=\textwidth]{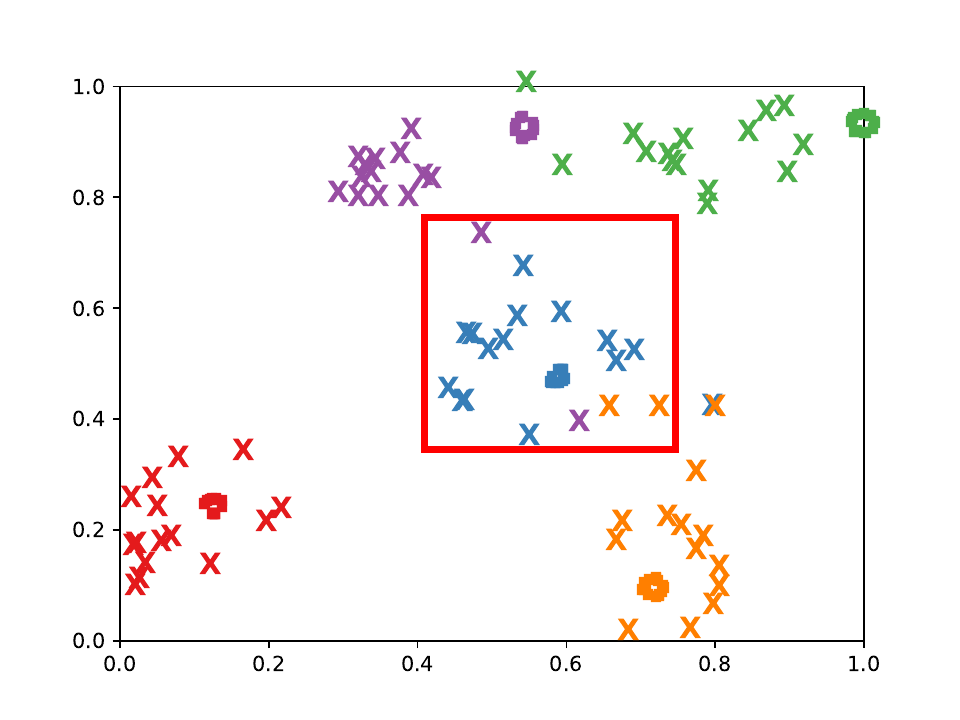}
     \caption{LWE on miniImageNet}
 \end{subfigure}
 
 \begin{subfigure}{0.32\linewidth}
     \includegraphics[width=\textwidth]{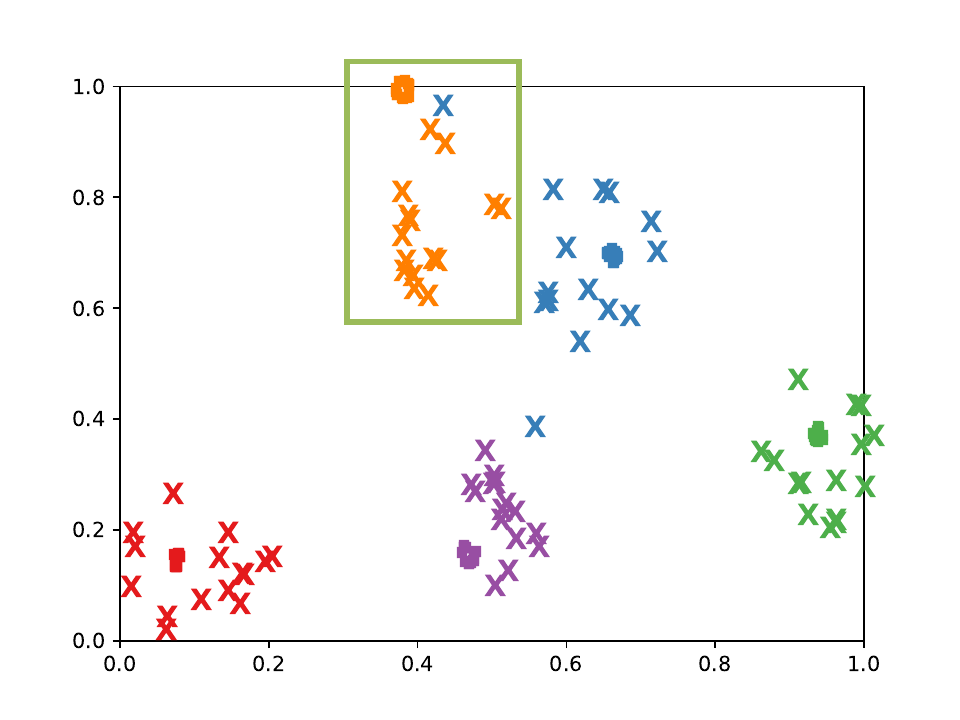}
     \caption{LWE + SPM on CIFAR-FS}
 \end{subfigure}
 \begin{subfigure}{0.32\linewidth}
     \includegraphics[width=\textwidth]{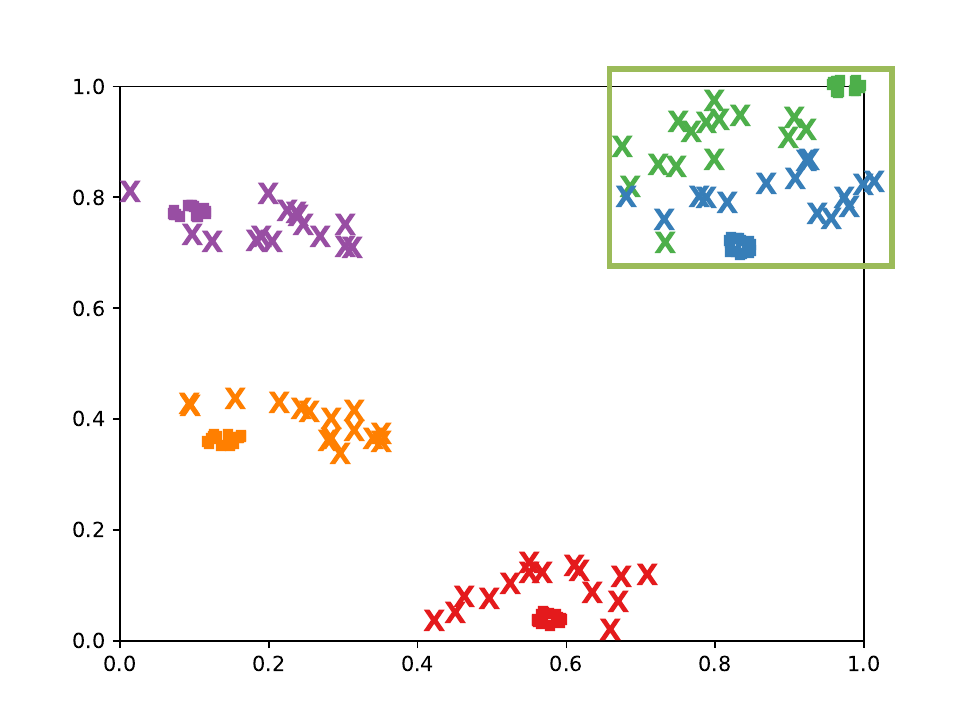}
     \caption{LWE + SPM on CUB-200-2011}
 \end{subfigure}
 \begin{subfigure}{0.32\linewidth}
     \includegraphics[width=\textwidth]{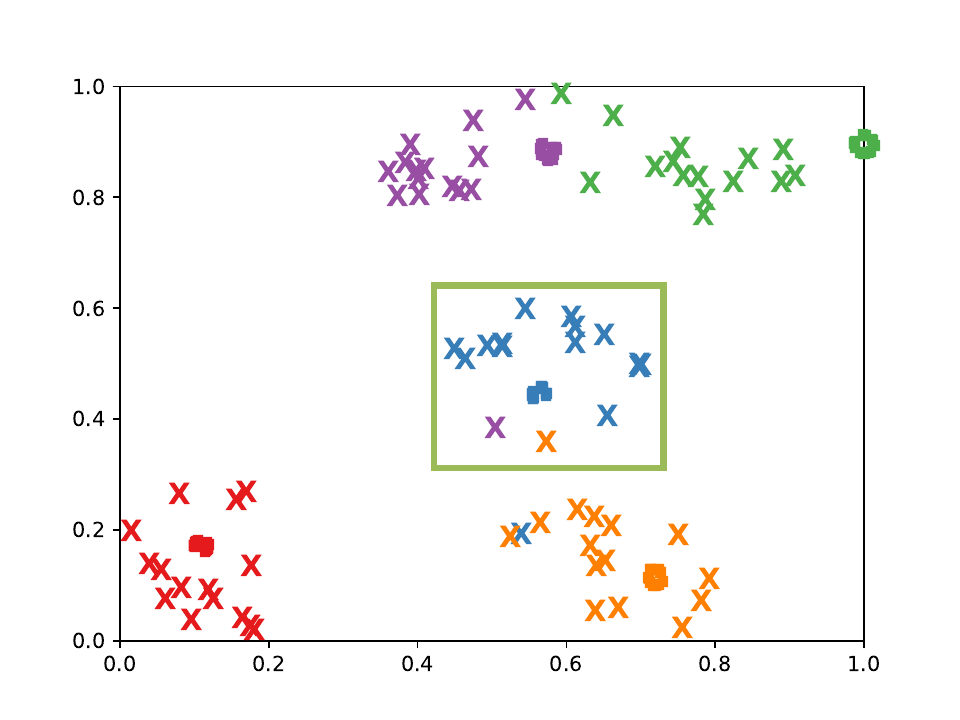}
     \caption{LWE + SPM on miniImageNet}
 \end{subfigure}
 \caption{The distribution of support and query set embeddings in the 5-way 1-shot task. ''·'' represents the support set and ''X'' represents the query set. With the integration of SPM, the query set gets closer to the support set, distinctions between different classes become more pronounced, and misclassifications are reduced.}
 \label{fig4}
\end{figure*}

Next, we conducted experiments on miniImageNet, tieredImageNet, CUB-200-2011, and CIFAR-FS to demonstrate the effectiveness of LWE and SPM. We selected ResNet18 as the baseline for comparison, with the results shown in Table \ref{t8}. Additionally, we employed t-SNE for dimensionality reduction and visualized the distribution of the final support and query embeddings, as illustrated in Figure \ref{fig4}. It can be observed that after incorporating the SPM module, the distance between the query set "·" and the support set "X" becomes smaller, and misclassifications are also less frequent. This demonstrates that by spatially aligning semantic information, our SPM module provides a more accurate measure of similarity between image pairs.

\textit{Layer-wise Embedding}. We choose the last two layers of the backbone as the layer-wise embedding. As shown in Table \ref{t8}, using LWE alone achieves performance very similar to that of MCNet. Our model achieves this performance without relying on any additional convolution or attention operations beyond the backbone, highlighting the effectiveness of LWE.

\textit{Semantic Pixel Matching}. Unlike MCNet, which uses multi-layer self-attention to learn the relationships between different correlation maps, our method employs the Hungarian algorithm to match the semantic pixels between query and support embeddings. This lightweight approach not only simplifies the process but also achieves superior performance.

Finally, we evaluated the scalability of our method on deeper backbones, as shown in Table \ref{t9}. Specifically, we used ResNet34 and ResNet50 as backbones for experiments on both MCNet and our method. For our approach, we followed the setup used with ResNet18, utilizing the outputs of the last two basic blocks. The results indicate that MCNet provides only a slight improvement over using the backbone alone, whereas our method achieves an obvious performance increase compared to both the backbone alone and MCNet. This difference arises because MCNet compresses multi-level correlation maps via an attention mechanism, effectively reducing redundancy on shallower backbones. However, with deeper backbones, it becomes challenging to capture the richer multi-level semantic information through a single-layer representation. In contrast, our Layer-wise Embedding retains multi-level semantic details and utilizes the Hungarian algorithm for refined feature alignment, enabling superior performance with deeper backbones. This demonstrates that our method can be extended to more powerful backbones to achieve additional performance gains.
\section{Conclusion}
In this paper, we proposed the Layer-wise Feature Metric of Semantic-Pixel Matching (LWFM-SPM) to address key challenges in few-shot learning. Our method efficiently integrates multi-level semantic features through the Layer-wise Embedding Module, ensuring a comprehensive use of feature information across different levels. Furthermore, we introduced the Semantic-Pixel Matching Module, which applies the Hungarian algorithm and a learnable matcher to handle semantic pixel misalignment, resulting in more accurate similarity scores between image pairs. Extensive experiments on benchmark datasets demonstrated the effectiveness and lightweight nature of our approach, achieving state-of-the-art performance.

Moving forward, we aim to explore further optimizations for more complex visual tasks. We plan to extend our method to handle other challenging scenarios such as domain adaptation and cross-modal few-shot learning, which could benefit from the principles established in this work. 

\bibliographystyle{elsarticle-num} 
\bibliography{references}

\end{document}